\documentclass[letterpaper, 10 pt, conference]{ieeeconf}  

\IEEEoverridecommandlockouts                              

\usepackage[textsize=footnotesize]{todonotes}
\usepackage[hidelinks]{hyperref}
\usepackage{url}
\usepackage{booktabs}
\usepackage{balance}

\title{\LARGE \bf
Investigation of Multiple Resource Theory Design Principles on Robot Teleoperation and Workload Management
}

\author{Zhao Han$^{1}$, Adam Norton$^{1}$, Eric McCann$^{1}$, Lisa Baraniecki$^{2}$, Will Ober$^{3}$,\\Dave Shane$^{3}$, Anna Skinner$^{2}$, and Holly A. Yanco$^{1}$
\thanks{$^{1\ }$University of Massachusetts Lowell, 1 University Avenue, Lowell, MA 01854. {\tt\small \{zhan,anorton,emccann,holly\}@cs.uml.edu}}%
\thanks{$^{2\ }$AnthroTronix, Inc., 8737 Colesville Rd, Suite L203, Silver Spring, MD 20910. {\tt\small \{baraniecki16,anna.d.skinner\}@gmail.com}}%
\thanks{$^{3\ }$Boston Engineering Corp., 300 Bear Hill Rd, Waltham, MA 02451. {\tt\small ober.will@gmail.com,
dshane@boston-engineering.com}}}%

\begin{document}

\maketitle
\thispagestyle{empty}
\pagestyle{empty}


\begin{abstract}

Robot interfaces often only use the visual channel. Inspired by Wickens' Multiple Resource Theory, we investigated if the addition of audio elements would reduce cognitive workload and improve performance. Specifically, we designed a search and threat-defusal task (primary) with a memory test task (secondary). Eleven participants -- predominantly first responders -- were recruited to control a robot to clear all threats in a combination of four conditions of primary and secondary tasks in visual and auditory channels. While we did not find any statistically significant differences in performance or workload across subjects, making it questionable that Multiple Resource Theory could shorten longer-term task completion time and reduce workload.
Our results suggest that considering individual differences for splitting interface modalities across multiple channels requires further investigation.

\end{abstract}


\section{Introduction}

Many robot control interfaces available to first responders for urban search and rescue (USAR) and counter-IED (improvised explosive device) scenarios  (e.g., \cite{yanco2007rescuing,larochelle2012multi,cichon2017robotic}) exclusively use the operators' visual channel to convey task-relevant information.
Such interfaces often require operators to prioritize their visual attention carefully, where information may be overloaded \cite{yanco2004}. By splitting the output of an interface across multiple sensory channels, the human-robot interface of a system could potentially foster appropriate attention management, thereby reducing cognitive workload, and improving situation awareness \cite{endsley2016designing} and performance.

To investigate this theory, we performed a within-subjects experiment in which participants, predominantly first responders, performed a search and (simulated) threat-defusal task (primary task) while simultaneously performing a periodic working memory task (secondary task). 
Each participant performed this task using an interface with four conditions of variable output modalities that were conveyed in the visual channel and/or the auditory channel.

\begin{figure}[t]
\centering
\includegraphics[width=\columnwidth]{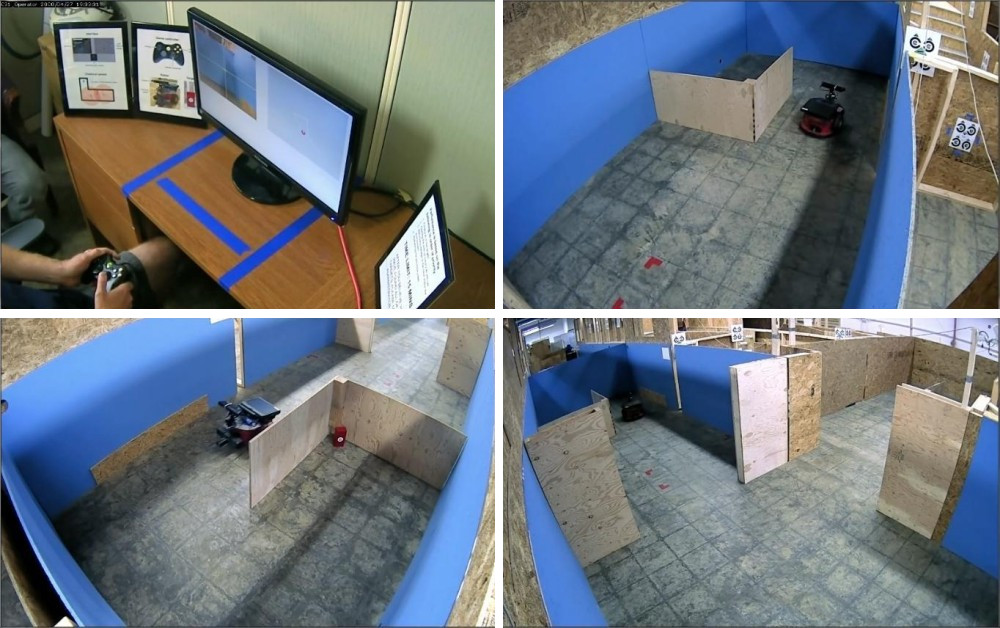}
\caption{To investigate whether a combination of audio and visual communication channels reduce workload and improve performance, we set up a test course and had a graphical interface (top left; also see Fig. \ref{fig:interface}) for robot operators to control a robot for a search and threat-defusal task (red block as threat, see bottom left). For this full run, please watch the accompanying video \cite{video}.}
\label{fig:multi-view}
\end{figure}

\section{Related Work}
  
Wickens' Multiple Resource Theory \cite{wickens1992} states that a cross-modal interface (using modalities that reside in two different channels) has advantages over an intra-modal interface (using modalities that reside in the same channel), as they use different resources in the brain. However, they can compete for common resources if both use the same ``code of processing,'' meaning each either conveys spatial data or verbal/linguistic data, taxing spatial versus verbal working memory, respectively. 

We designed our interfaces based on experimentally-informed modifications of interfaces from similar user studies \cite{bkeys2010, Desai:2012-phdthesis, micire2011design}.
Design guidelines for robot control interfaces were also leveraged, such as the use of sensor fusion to reduce cognitive load on the operator~\cite{yanco2004}.

Krausman et al. \cite{krausman2005effects} conducted an experiment to determine, of visual, auditory, and tactile cues, which was a more effective channel through which to alert platoon leaders for decision-making. By comparing the amount of time it took to react to an alert, results showed that auditory alerts resulted in the fastest response time. Visual alerts were the slowest. Other researchers had similar findings \cite{folds1994auditory,dixon2005mission,prewett2010managing}. Given that the majority of an interface's output modalities are in the visual channel, it suggests that we should offload some visual information to the audio channel. For the combination of the three cues, Benz and Nitsch~\cite{benz2017using} found that using all of them deteriorates performance. In this work, we use only visual and auditory cues, focusing on overall task performance instead of short-term response times.

In the field of adaptive automation, there have been studies related to the performance and workload implications of varying the modality of cues to indicate dynamic-automation state changes \cite{Kaber2006527}. In a simulated mine-disposal task, Kaber et al. found that bimodal cueing of autonomy state changes significantly reduced task completion time, but did not find a significant effect on workload between visual and multi-modal cueing conditions.

\section{Experiment Design}

\begin{figure}[t]
\centering
\includegraphics[height=1.3in]{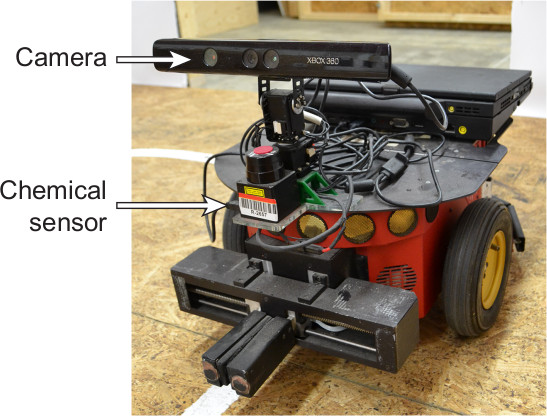}\hspace{0.2in}
\includegraphics[height=1.3in]{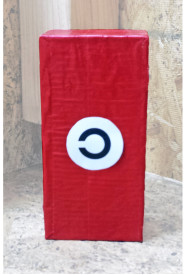}
\caption{\underline{Left}: The Adept MobileRobots Pioneer remote-controlled by participants. \underline{Right}: A ``threat'' in the test course. The threat is removed if the correct direction of the Landolt C is read (left in this example).}
\label{fig:robot_and_threat}
\end{figure}

\subsection{Robot Platform}

An Adept MobileRobots Pioneer, equipped with a Hokuyo URG laser rangefinder and a Microsoft Kinect RGBD camera mounted to a pair of Dynamixel RX-28 servos acting as the robot's camera and pan-tilt unit (Fig. \ref{fig:robot_and_threat}, left) was used in the experiments.
Participants were told that the robot was equipped with a chemical sensor to locate threats in the area that would be conveyed through a targeting system; however, this sensor was simulated using the robot's knowledge of where the threats were located on the map.
ROS (Robot Operating System) \cite{ros} was used to control the robot, communicate between the robot and the interface, and log performance metrics during task execution.

\subsection{Tasks}

This experiment sought to verify the relationships between information modalities and cognitive workload as presented in Multiple Resource Theory.
A \textit{targeting system} (more in Section \ref{sec:interface}), to aid in the execution of the primary task and the secondary task instructions, was varied between visual and auditory channels.

\begin{figure}[t]
\centering
\includegraphics[width=0.7\columnwidth]{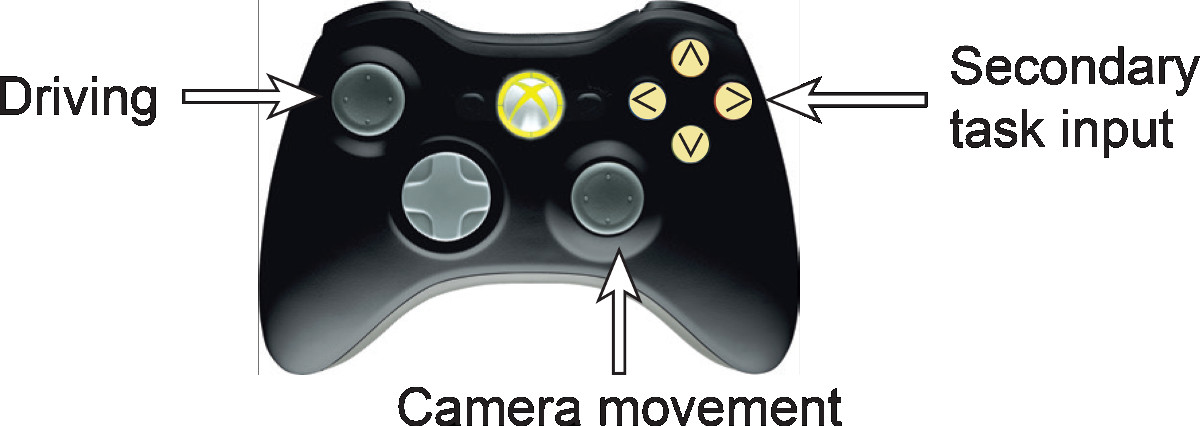}
\caption{The game controller used by participants to control the robot in both the primary and secondary tasks.}
\label{fig:controller}
\end{figure}

\subsubsection{Primary Task} The primary task is a search and threat-defusal task where participants need to drive the robot to clear all of the threats. We set up a $\sim92 m^2$ test course (Fig. \ref{fig:multi-view}) that consisted of rooms, hallways, and occlusions, with four threats hidden throughout.

The occlusions were designed such that the participant could only see a threat in the robot's camera after having driven behind the occlusion.
This was done to ensure that rooms and occlusions had to be investigated by participants, rather than allowing the participant to see a threat from the hallway.
It was also aimed to influence participants to rely on the targeting system to search for threats, otherwise they risk wasting time investigating empty occlusions.

Participants drove the robot by controlling its linear and rotational velocities with the left joystick of a Microsoft Xbox 360 controller (Fig. \ref{fig:controller}). The right joystick was used to pan/tilt the robot's camera for different views.

\begin{figure}[t]
\centering
\includegraphics[width=0.9\columnwidth]{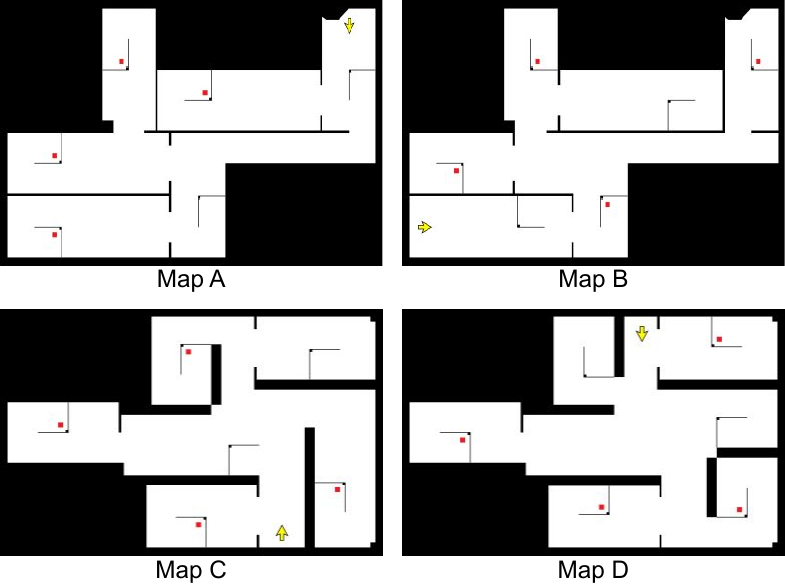}
\caption{The four maps used for the search and threat-defusal task. Maps A and B were of similar shape, as were Maps C and D. The initial position of the robot is indicated with a yellow arrow and threat locations are the four red boxes in each map.
}
\label{fig:maps}
\end{figure}

\subsubsection{Secondary Task} While the targeting system could affect an operator's workload, participants could ignore the system and, for example, perform a right-hand wall follow\footnote{\url{https://en.wikipedia.org/wiki/Maze_solving_algorithm\#Wall_follower}} -- a common practice of first responders -- while still performing the primary task effectively. For this reason, we implemented a secondary task, to induce an additional cognitive workload on participants in a way that would not vary with the individual participant's primary task strategy.

The secondary task is a periodic working memory task, consisting of a random sequence of five commands selected from four choices (up, down, left, and right) that the system would convey to the operator (see Fig. \ref{fig:interface} for an example). Once the system completed the sequence with the word ``over'', the participant would repeat the five commands using the game controller's directional pad (Fig. \ref{fig:controller}).
Essentially, the task was a linguistic casting of Milton Bradley's Simon\footnote{\url{http://en.wikipedia.org/wiki/Simon_(game)}}.

Each input by the operator elicited feedback from the system.
If a direction was pressed before ``over'' occurred, ``early'' was indicated.
Correct and incorrect individual steps were indicated with either positive or negative feedback.
Once the entire sequence was correctly repeated, the system would indicate ``done''.
If incorrect, the operator needed to correctly enter it before entering the next direction in the sequence.
Tasks with unresponded steps were counted as incomplete when the next task's instructions began, 30 seconds from the beginning of the previous instructions.
 
\subsection{Control Interface}\label{sec:interface}

\begin{figure}[t]
\centering
\includegraphics[width=\columnwidth]{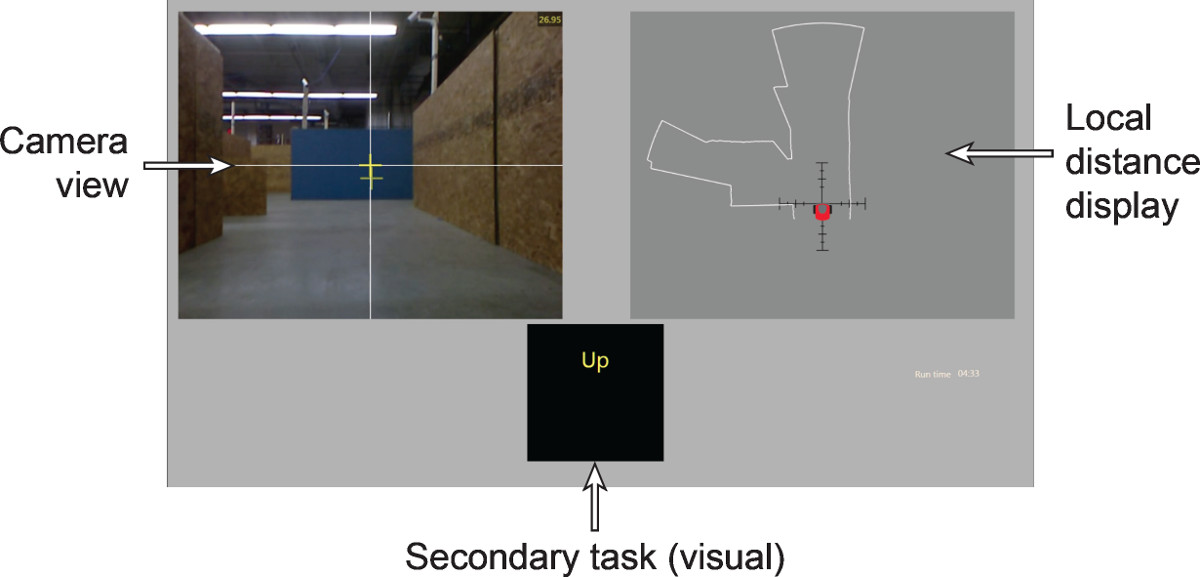}
\caption{The graphical control interface showing the camera view (left) and local distance display (right). A directional command in the secondary task is shown at bottom; The direction text is hidden when conveyed with audio.
In the visual targeting system, when a threat is sensed, a yellow line connects the red robot avatar in the top right to the threat.
}
\label{fig:interface}
\end{figure}

The interface consisted of two main output modalities: a camera view with crosshair and a local distance display (Fig. \ref{fig:interface}). The crosshair represents the pan-tilt position of the camera. The local distance display shows a two-dimensional top-down view of the boundaries within 270 degrees around a local avatar of the robot; no local distance data was available directly behind the robot.
Input was provided using a gamepad (Fig. \ref{fig:controller}).
The position, size, and behavior of the main output modalities were constant across all conditions.

To help participants find threats, a simulated targeting system was integrated either into the graphical interface or through sound.
To clear a threat, participants needed to correctly read the direction of a Landolt C that was on its side (Fig. \ref{fig:robot_and_threat} right).
After a threat was removed, the targeting system would no longer detect it.

The targeting system sensed the closest threat within 2.4 meters and 180 degrees in front of the robot. It presented the detected threat's position two-dimensionally.
This distance threshold was intentionally chosen to require participants to enter a room to determine if a threat was present.

The visual targeting system took the form of a yellow line overlaid on the local distance display, connecting the robot avatar to the sensed threat.
The visual targeting system was the only information that was colocated with other visual information.
Like the local distance display, the visual targeting information was presented as a top-down orthographic projection, and both shared a rotational origin.
To prevent the visual targeting from being interpreted as communicating an absolute distance, which the auditory targeting system could not provide, the visual targeting system's line length was constant.
Its thickness varied with distance to the threat.

The auditory targeting system was heard through stereo headphones worn by participants, and was implemented using OpenAL's 3D spacialization capability. The sound sources were restricted to the horizontal plane through the origin.
This audio signal presented a threat's location as a periodic ping in the direction of the threat from the robot's front, at a distance directly related to the threat's distance.

The visual secondary task conveyed commands to the operator by displaying the directions in word form, centered below the main visual modalities.
Visual secondary task feedback was presented instructions in the same region, but on a line below.
The words ``Yes'' and ``No'' were presented as feedback for correct and non-early incorrect responses.

The auditory modality in the secondary task conveyed commands and all responses to the operator with text-to-speech.
Correct and non-early incorrect response feedback was presented as a positive ding and a negative buzzer sound, respectively.
The sounds and their meaning were presented to participants in the secondary task introduction.
``Early'' feedback was presented at a lower volume and faster speech rate than instructions, so that the instruction timing would not be changed by feedback for an early response.

\subsection{Procedure}
For this within-subjects experiment, the search task was performed four times by each participant, covering all four combinations of the two independent variables -- the modality in which the targeting system conveyed proximal threat locations (visual or auditory) and the modality in which the secondary task instruction was conveyed (visual or auditory) -- counterbalanced to mitigate any learning effects.

Each run was performed in one of four maps (Fig. \ref{fig:maps}); ordering was also counterbalanced to mitigate any learning effects. Every map contained five rooms (shown with thick lines), a hallway that connected them (center, 1.2$m$ tall walls), and six occlusions (thin lines, 0.6$m$ tall walls).
Four of the occlusions had a threat (red square) behind them.

Participants were not told how many threats were on each map.
They were instructed to announce the end of their run if they felt they had completed their search and found all possible threats.
Otherwise, they had a time limit of 15 minutes.
They were also not told how often the secondary task would occur during each run.

Based on real-world search and threat-defusal scenarios, participants were told their priorities for task execution were:

\begin{enumerate}
\item Finding all of the threats
\item Clearing the area as quickly as possible
\item Minimizing damage to the robot and the environment
\item Performing the secondary task
\end{enumerate}

First, the primary and secondary tasks were described to participants.
We also printed annotated photos and placed them on the desk as a reminder (Fig. \ref{fig:multi-view} upper left).
After a task was described, participants used each of the modalities for that task, separately, to ensure that participants were familiar with each of the variable modalities without the interference of another modality.
The primary task modalities were introduced on a simplified map.
The secondary task modalities were used on a simplified interface.
Participants were allowed to continue learning each task until they reported feeling comfortable with it.

This study was approved by the Institutional Review Board (IRB) at the University of Massachusetts Lowell.

\subsection{Conditions}\label{sec:conditions}

Each combination of auditory and visual modalities produced four interface conditions, one used for each participant's four runs (abbreviated, PrimarySecondary):
\begin{enumerate}
\item \textbf{AV}: auditory targeting and visual secondary task 
\item \textbf{VA}: visual targeting and auditory secondary task 
\item \textbf{AA}: auditory targeting and auditory secondary task 
\item \textbf{VV}: visual targeting and  visual secondary task 
\end{enumerate}

\subsection{Hypotheses}

According to Multiple Resource Theory, heterogeneous modalities (AV and VA) should have a lower workload and higher performance than conditions with homogeneous modalities (AA and VV).
Therefore, we hypothesized:

\textbf{Hypothesis 1 (H1).} Conditions with heterogeneous modalities (AV and VA) should result in a lower cognitive workload than conditions with homogeneous modalities (AA and VV).
Lower cognitive workload is measured by subjective measures in Table \ref{tab:task-load-index} and data in the secondary memory task: more fully correct command sequences inputted; lower percentage of incorrect inputs; and subjective measures such as stress and mental demand.

\textbf{Hypothesis 2 (H2).} Conditions with heterogeneous modalities (AV and VA) should result in higher performance than conditions with homogeneous modalities (AA and VV).
Higher performance is measured by the primary task data:
less time to clear the area; less time to find each threat; more threats found; and fewer bumps and collisions.

Because the visual targeting system and local distance display were fused, which should reduce operator workload \cite{yanco2004}, two additional hypotheses were formed:

\textbf{Hypothesis 3 (H3).} Condition VA should result in lower cognitive workload than AV, using the same measures as~H1.

\textbf{Hypothesis 4 (H4).} Condition VA should result in higher performance than AV, measured the same way as H2.

\subsection{Data Collection and Questionnaire}

\begin{table}
\caption{Adapted NASA-Task Load Index \cite{hart2006nasa} Questions}
\label{tab:task-load-index}
\begin{tabular}{p{0.9\columnwidth}}
\toprule
\textbf{Stress}: \textit{How stressful the task was}\\
\midrule
\textbf{Mental Demand}: \textit{How taxing the task was on you mentally}\\
\midrule
\textbf{Physical Demand}: \textit{How taxing the task was on you physically}\\
\midrule
\textbf{Effort}: \textit{How much effort you think you applied to complete the task}\\
\midrule
\textbf{Frustration}: \textit{How frustrated you were while completing the task}\\
\bottomrule
* Likert items are coded from 1 (Very Low) to 7 (Very High).
\end{tabular}
\end{table}

We were able to collect and compute most data directly from the robot and the control interface. However, to count the number of times the robot bumped into the environment, we set up a multi-angle camera system so experimenters could record the collisions during the experiment. Another camera recorded the operator with the interface.

After each run, participants were given 7-point Likert scale questions to get workload estimates, adapted from the NASA-Task Load Index \cite{hart2006nasa}. Table \ref{tab:task-load-index} lists the questions.

\subsection{Participants}

Thirteen participants were recruited. Two were involved in pilot testing, so 11 participants contributed valid data (1 Female, 10 Male); eight were first responders (73\%; 1 Female, 7 Male). Each participant was compensated US \$75 for participation and travel.  Age ranged from 20 to 55 ($M=40$, $SD=10$). All participants reported no hearing problems and their primary language was English.

In terms of game controller usage, 2 rated disagree, 2 reported neutral, and 7 agreed; we had a similar report for playing first-person perspective video games (2 vs. 3 vs. 6). When asked whether they have experience with operating robots, 4 disagreed, 2 rated neutral, and 5 agreed.

\section{Results}

\subsection{No detectable workload difference (H1 \& H3 rejected)}

\subsubsection{Correct command sequences}

\begin{figure}
\centering
\includegraphics{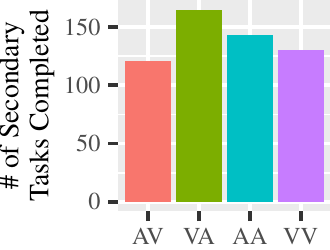}\hspace{0.3in}
\includegraphics{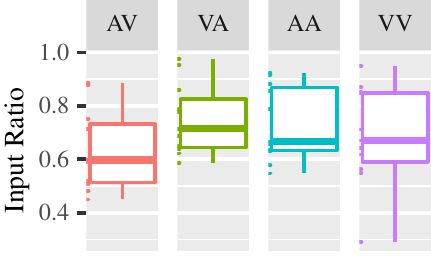}
\caption{\underline{Left}: Secondary task completion count for each condition ($p=0.05$). \underline{Right}: Secondary task correct input ratio ($n.s.$).}
\label{fig:SecondaryTask}
\end{figure}

We first performed a chi-square goodness-of-fit test on the frequency of correct command sequences in the secondary task across 4 conditions. The test did not reveal a statistical significance ($p=0.05$), thus we did not run a post-hoc pairwise comparison. The frequency data for each condition is in Fig. \ref{fig:SecondaryTask} (left).

\subsubsection{Incorrect inputs}

We then conducted a repeated-measures ANOVA on the correct input ratio from the secondary task. A box plot is shown in Fig. \ref{fig:SecondaryTask} (right).
Prior to the test, we ran 2 tests to assess the underlying assumptions of repeated-measures ANOVA: normality and sphericity. We ran the Shapiro-Wilk normality test; the result showed that the data is normally distributed ($W=1, p=0.8$). We ran Mauchly's test for sphericity when showed that sphericity is violated ($\chi^2(3)=0.12, p<0.005$).
We thus used a Greenhouse-Geisser correction. There was a significant difference across conditions ($F(1.756,17.562)=3.907, p=0.044$). However, after we conducted a Bonferroni-corrected pairwise comparison with paired-samples t-tests, we did not find any statistically significant differences, partially because the p-value $0.044$ is very close to $0.05$.

\subsubsection{NASA-Task Load Index responses}

\begin{figure}
\centering
\includegraphics{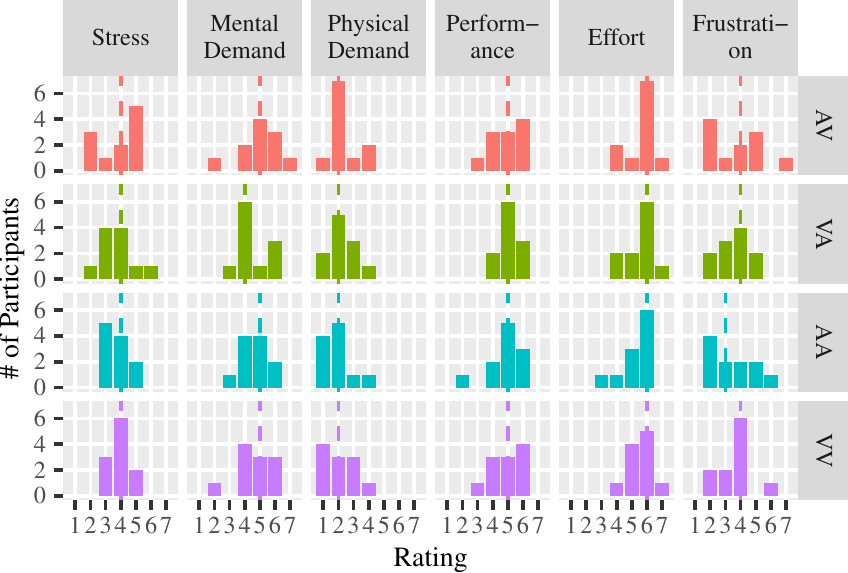}
\caption{All questionnaire responses. Dashed lines show median values ($n.s.$ between conditions).}
\label{fig:qbars}
\end{figure}

We analyzed participants' responses to the NASA-Task Load Index questionnaire (Table \ref{tab:task-load-index}). From the bar plot in Fig. \ref{fig:qbars}, there seemed to be no differences across questions and conditions. This is further confirmed by a non-parametric Friedman's test on each metric: Stress ($\chi^2(3)=0.27, p=1$), Mental Demand ($\chi^2(3)=4.1, p=0.2$), Physical Demand ($\chi^2(3)=7.1, p=0.07$), Performance ($\chi^2(3)=1.7, p=0.6$), Effort ($\chi^2(3)=1.6, p=0.7$), and Frustration ($\chi^2(3)=0.46, p=0.9$).

\subsection{No detectable performance difference (H2 \& H4 rejected)}

\subsubsection{Clearing the arena}

\begin{figure}
\centering
\includegraphics{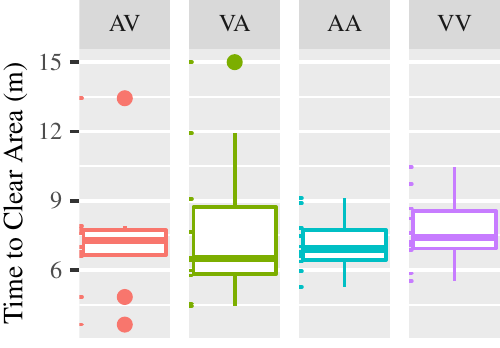}
\caption{Time to clear the area ($n.s.$)}
\label{fig:PriTaskCompletionTime}
\end{figure}

As shown in Fig. \ref{fig:PriTaskCompletionTime}, participants that experienced each condition finished clearing the threats in approximately 8 minutes on average.

We ran a repeated-measures ANOVA. Again, we ran tests for the underlying assumptions of ANOVA. The Shapiro-Wilk normality test only showed a slight violation of normality ($W=0.84, p = 0.043$), and ANOVA is generally considered robust against normality moderate violations \cite{pagano2012understanding}. Mauchly's test showed no violation for sphericity ($\chi^2(3)=0.456, n.s.$). There were no statistically significant results found ($F(3,11)=0.228, n.s.$).

\subsubsection{Finding each threat}

\begin{figure}
\centering
\includegraphics{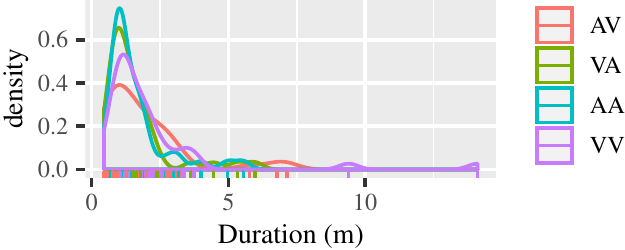}
\caption{Time to find each threat ($n.s.$). A density plot with rug plot is used due to more similar distributions across conditions.}
\label{fig:PriTime2FindEachThreat}
\end{figure}

As seen in Fig. \ref{fig:PriTime2FindEachThreat}, there were no statistically significant differences across conditions on time to find each threat, revealed by repeated-measures ANOVA with Greenhouse-Geisser correction ($F(1.83,18.28)=0.86, n.s.$). The Shapiro-Wilk test did not reveal a normality violation ($W=0.90, n.s.$), but Mauchly's test showed a violation of sphericity ($\chi^2(3)=0.25, p<0.05$).

\subsubsection{Total threats found}

\begin{figure}
\centering
\includegraphics{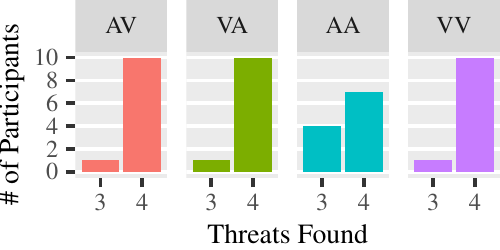}
\caption{The number of threats found ($n.s.$). Most participants found all.}
\label{fig:PriThreatFoundBar}
\end{figure}

We ran Friedman's test for the number of threats found across conditions. No statistically significant differences were found ($\chi^2(3)=5.4, p=0.14$). The median values for all conditions are 4 threats, indicating that most participants found all of the threats (Fig. \ref{fig:PriThreatFoundBar}).

\subsubsection{Bumps and collisions}

\begin{figure}
\centering
\includegraphics{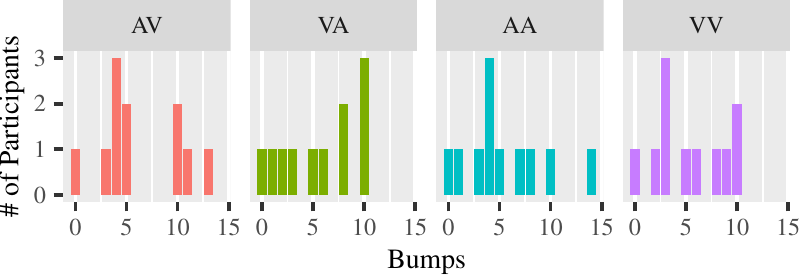}
\caption{The number of bumps and collisions ($n.s.$).}
\label{fig:PriBumps}
\end{figure}

We also ran Friedman's test to determine if there were any statistically significant differences for the number of bumps across conditions; it did not show any significance ($\chi^2(3)=1, p=0.7$). Fig. \ref{fig:PriBumps} shows the data in bar plots.

\subsection{Individual Differences}

Inspired by the lower-body exoskeleton work by Blake \textit{et al}. \cite{bequette2020physical} on physical and cognitive load effects, we investigated the differences in performance within individuals. A better understanding of individual variability helps understand a wider set of the population~\cite{bequette2020physical}, so we treated the participant as a factor and explored the data. We excluded the data with only 1 data point per participant. Because the time to find a threat is not a balanced factor, for its test, where the normality violation was significant, Friedman's test could not be conducted.

\begin{figure}
\centering
\includegraphics{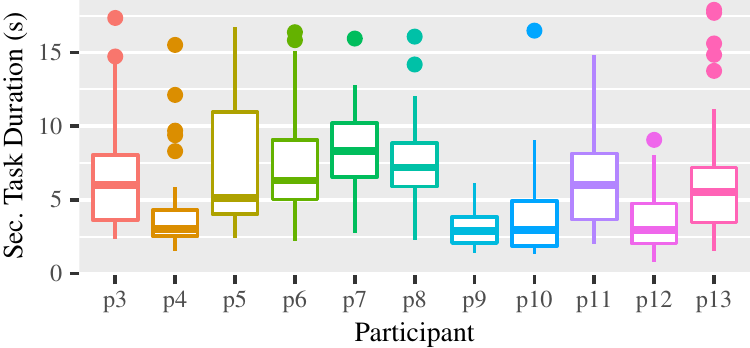}
\caption{Time to input a correct command sequence.
}
\label{fig:SecTaskTimeByParticipants}
\end{figure}

\begin{figure*}
\centering
\includegraphics{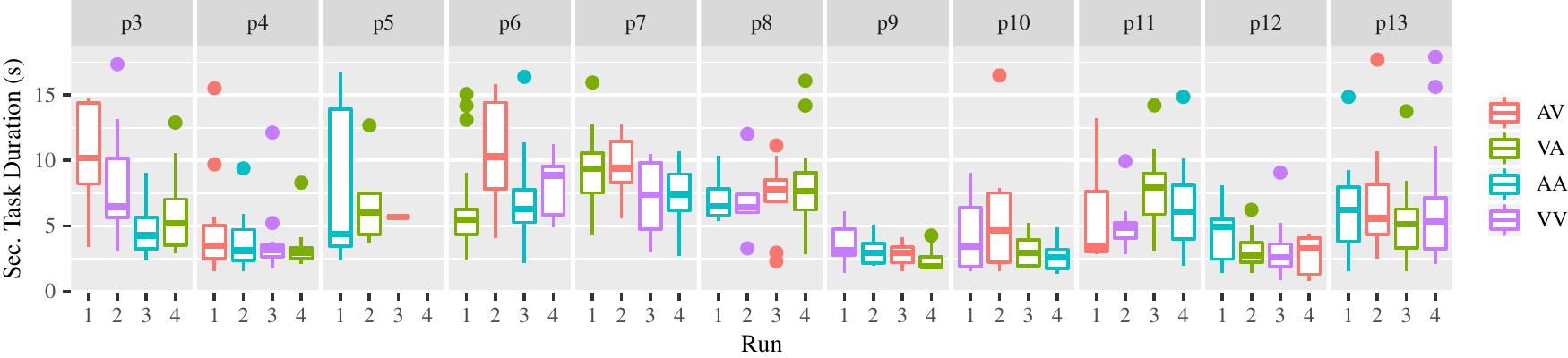}
\caption{Time to input a correct command sequence, grouped by participant and the order of runs.}
\label{fig:SecTaskTimeByParticipantCondition:pagewidth}
\end{figure*}

We found that there are statistically significant differences \textit{between} 
most participants in the time to input a correct command sequence during the secondary task (Fig. \ref{fig:SecTaskTimeByParticipants}). Due to a significant violation of normality by Shapiro-Wilk test ($W=0.93, p<0.0005$), this is determined by Kruskal-Wallis H tests ($\chi^2(10)=188.49, p<0.0005$) and pairwise Wilcoxon tests with adjusted p-values using Benjamini and Hochberg (BH) methods \cite{benjamini1995controlling} (see \cite{wilcoxonparticipant} for the full list; as a rule of thumb, $p<0.05$ in Fig. \ref{fig:SecTaskTimeByParticipants} if no significant overlap).

Upon further exploration for each participant, we found there were effects \textit{within} participants p3, p6, and p7 across condition ordering in the time to input a correct command sequence (Fig. \ref{fig:SecTaskTimeByParticipantCondition:pagewidth}). When there is significant violation of normality by Shapiro-Wilk test, we ran Kruskal-Wallis H tests; otherwise one-way ANOVA was used. The 3 statistically significant results were determined by Kruskal-Wallis H tests (p3: $\chi^2(3)=8.595, p<0.05$; p6: $\chi^2(3)=11.132, p<0.05$; p7: $\chi^2(3)=10.619, p<0.05$). However, pairwise Wilcoxon tests with adjusted p-values using Benjamini and Hochberg (BH) methods \cite{benjamini1995controlling} only revealed one statistically significant difference ($p<0.05$) between p6's run 1 (Median=$5.46s$) and run 2 (Median=$10.28s$).
This variability within and between operators suggests that there may be benefits to the development of adaptable interfaces, which could adjust to individual capabilities and preferences.

\section{Discussion}

Surprisingly, no detectable differences between conditions were found for all 7 measures. We did not find statistical evidence showing that heterogeneous modalities (AV, VA) have lower workload and higher performance than conditions with homogeneous modalities (AA, VV).

Our study casts doubt on using Multiple Resource Theory as a design principle to improve performance by splitting information into multiple channels, when utilized for more regularly occurring information in a longer-term interaction scenario. If only utilized for the presentation of less frequent alerts, as supported by the findings of previous research (e.g., \cite{krausman2005effects,folds1994auditory,dixon2005mission,prewett2010managing}) that audio cues make response time faster, it may prove to be a more fruitful design principle, but further investigation is needed.

We did find statistically significant evidence of individual differences (Fig. \ref{fig:SecTaskTimeByParticipants}).
The impact of individual differences on performance when using robot interfaces has been studied in similar domains (e.g.,~\cite{barnes2011designing,chen2011individual,chen2012supervisory}, which typically involve comparing an operator's performance on pre-test attentional and spatial ability tasks to those involving control of one or more robots.
Such characterizations of individual differences can predict the performance of operators when using robots with interfaces using only visual modalities.

Some participants showed general improvements in secondary task performance when it was deployed in a particular modality (Fig. \ref{fig:SecTaskTimeByParticipantCondition:pagewidth}).
For example, p3, p6, and p10 typically performed the secondary task faster when it was presented in the audio channel (VA, AA) and p5 didn't respond to the secondary task at all during the VV condition, possibly because they were overloaded.
One participant, p11, typically performed better when the secondary task was presented in the visual channel (AV, VV).
Participants such as p4, p8, p9, p12, and p13 appeared to perform similarly regardless of how the secondary task was presented.
Further investigation is needed to determine if these individual differences are worth characterizing to tailor interface presentation per each operator's capabilities and preferences.

The effect of individual differences should be further investigated as to its implications on how and when to implement Multiple Resource Theory design principles on a robot control interface.
Pre-test characterizations could be used to tailor an interface before usage. 
Performance also may degrade over time throughout task duration, which could call for an adaptive interface that recognizes this degradation and augments how the interface modalities function.
For example, if an interface begins in the VV condition, but secondary task performance degrades, it would adapt to the VA condition.
This would follow Multiple Resource Theory as a means to improve performance when necessary rather than as an overarching design principle which, as demonstrated by this study, may not be the appropriate method of implementation for the search and rescue domain or more broadly for longer-term interaction scenarios.

\section*{Acknowledgements}

This work has been supported in part by DARPA under W31P4Q-13-C-0136 and W31P4Q-13-C-0196; Distribution Statement A Approved for Public Release, Distribution Unlimited.
The views, opinions, and/or findings contained in this article are those of the authors and should not be interpreted as representing the official views or policies of the Department of Defense or the U.S. Government. 


\balance
\bibliographystyle{IEEEtran}
\bibliography{bib}

\end{document}